\documentclass[runningheads]{llncs}
\usepackage{graphicx}
\def\onedot{. }
 
\def\ie{\emph{i.e}\onedot}

\def\wrt{w.r.t\onedot}

\usepackage{algorithm}
\usepackage[noend]{algpseudocode}
\makeatletter
\def\BState{\State\hskip-\ALG@thistlm}
\makeatother

\usepackage{epsfig}
\usepackage{amsmath}
\usepackage{amssymb}
\usepackage{booktabs}

\begin{document}
\title{Non-Parametric Contextual Relationship Learning for Semantic Video Object Segmentation}
\titlerunning{Non-Parametric Contextual Relationship Learning for Semantic Video Object Segmentation}
%
\author{Tinghuai Wang\inst{1} \and
Huiling Wang\inst{2}\thanks{Corresponding author}}
%
%
\institute{Nokia Technologies, Tampere, Finland \\
	\email{tinghuai.wang@nokia.com} \and
Laboratory of Signal Processing, Tampere University of Technology, Finland \\
\email{huiling.wang@tut.fi}}
\maketitle              
\begin{abstract}
	We propose a novel approach for modeling semantic contextual relationships in videos. This graph-based model enables the learning and propagation of higher-level spatial-temporal contexts to facilitate the semantic labeling of local regions. We introduce an exemplar-based nonparametric view of contextual cues, where the inherent relationships implied by object hypotheses 
are encoded on a similarity graph of regions. Contextual relationships learning and propagation are performed to estimate the pairwise contexts between all pairs of unlabeled local regions. Our algorithm
integrates the learned contexts into a Conditional Random Field (CRF) in the form of pairwise potentials and infers the per-region semantic labels. We evaluate our approach on the challenging YouTube-Objects dataset which shows that the proposed contextual relationship model outperforms the state-of-the-art methods.

\end{abstract}
\section{Introduction}

Semantic object segmentation in videos is a challenging task which enables a wide range of higher-level applications, such as robotic vision, object tracking, video retrieval and scene understanding. 
Tremendous progress has been witnessed lately toward this problem via integrating higher-level semantic information and contextual cues \cite{HartmannGHTKMVERS12,TangSY013,taylor2013semantic,LiuTSRCB14,ZhangCLWX15,wang2016semi,drayer2016object,wang2017}. However, akin to classical figure-ground video segmentation, fast motion, appearance variations, pose change, and occlusions pose significant challenges to delineate semantic objects from video sequence. Difficulty in resolving the inherent semantic ambiguities further complicates the problem.

Recently, segmentation by detection and tracking approaches have been proposed to address this challenging problem. Early work in this direction trained classifiers to incorporate scene topology and semantics into pixel-level object detection and localization \cite{taylor2013semantic}. Later, both object detector and tracker were employed to either impose spatio-temporal coherence \cite{ZhangCLWX15,drayer2016object} or learn an appearance model \cite{wang2016semi} for encoding the appearance variation of semantic objects. Lately, hierarchical graphical model has also been proposed to integrate longer-range object reasoning with superpixel labeling \cite{wang2017}. Despite of significant advances that have been made by the above methods, global contextual relationships between semantic video objects remain under-explored. Yet, contextual relationships are ubiquitous and provide important
cues for scene understanding related tasks. 

The importance of exploiting pairwise relationships between objects has been highlighted in semantic segmentation \cite{lin2016efficient} and object detection \cite{gidaris2015object} tasks, where the relationship is formulated in terms of co-occurrence of higher-level statistics of object class. These methods tend to favor frequently appeared objects in the training data to enforce rigid semantic label agreement. Furthermore, these conventional context models are sensitive to the number of pixels or regions that objects occupy, with one consequence being that the small objects are more likely to be omitted. 

Graphical models have emerged as powerful tools in modeling  contextual relationships in computer vision \cite{wang2010multi,qi20173d,wang2017submodular,wang2019zero,wang2019graph,yang2021learning}, offering a versatile framework for representing and analyzing complex visual data. These approaches leverage the inherent structure and relationships within images \cite{wang2015robust,tinghuai2016method,xing2021learning} and videos \cite{wang2010video,wang2014wide,chen2020fine}, enabling more sophisticated and context-aware analysis. By representing visual elements as nodes and their interactions as edges, graphical models can capture spatial, temporal, and semantic dependencies crucial for various tasks such as visual information retrieval \cite{hu2013markov}, stylization \cite{WangCSCG10,wang2011stylized,wang2013learnable}, object detection \cite{WangW16,tinghuai2016apparatus,zhao2021graphfpn}, scene understanding \cite{WangW14,wang2017cross,wang2020spectral,tinghuai2020watermark,deng2021generative}, and image or video segmentation \cite{wang2015weakly,wang2016semi,wang2016primary,tinghuai2017method,tinghuai2018method1,ZhuWAK19,zhu2019cross,tinghuai2020semantic,lu2020video,wang2021end}.

In this work, we propose a novel graphical model to thoroughly exploit contextual relationships among semantic video objects without relying on training data. Such a way of modeling spatio-temporal object contextual relationships has not been well studied.  We present a novel nonparametric approach to capture the intra- and inter- category contextual relationships by considering the content of an input video. This nonparametric context model is comprised of a set of spatial-temporal context exemplars via performing higher-level video analysis, \ie object detection and tracking. These context exemplars provide a novel interpretation of contextual relationships in a link view which formulates the problem of learning contextual relationships as the label propagation problem on a similarity graph. This similarity graph naturally reflects the intrinsic and extrinsic relationship between semantic objects in the spatial-temporal domain. Due to the sparsity of this similarity graph, the learning process can be very efficient. 

The key contributions of this work are as follows. Firstly, we establishes a novel link prediction view of semantic contexts. In this view, the problem of learning semantic relationships is formulated as graph-based label propagation problem. Secondly, our approach is exemplar-based nonparametric model which therefore does not require additional training data to build an explicit context model. Hence, it is favorable for video semantic object segmentation, a domain where annotated data are scarce. The paper is organized as follows. We introduce the novel link predition view of contexts in Sec. \ref{sec:modeling}, utilizing the semantic contextual information from object trajectory hypotheses in Sec. \ref{sec:proposal}. Link prediction algorithm is described in Sec. \ref{sec:prediction} and the final semantic labeling is described in Sec. \ref{sec:inference}.

\begin{figure*}[t!]
	\centering
	\includegraphics[width=0.99\linewidth]{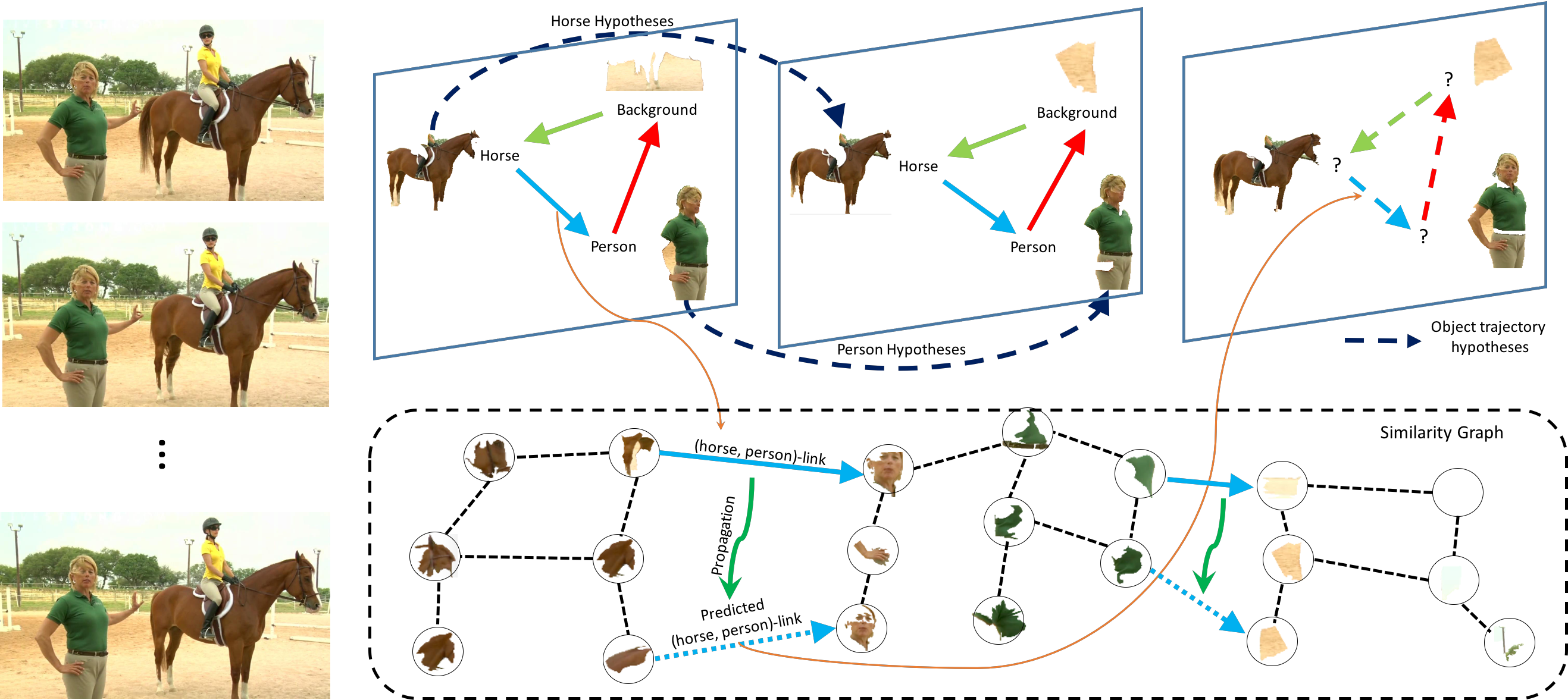}
	\caption{Illustration of the proposed approach. Two trajectory hypotheses are extracted to provide initial annotations for `horse' and 'person' classes, which form the context exemplars such as  (horse, person)-link between the corresponding vertices on the similarity graph. Our method propagates such contextual relationship on the graph and predicts the probability of (horse, person)-link between unlabeled vertices based on similarity. \label{fig:diagram}}
\end{figure*}

\section{The Approach}

In this section, we describe our proposed exemplar-based nonparametric model and how the learned contextual relationships are integrated into semantic labeling in a principled manner.

\subsection{Trajectory Hypotheses}
\label{sec:proposal}
For a given video sequence with T frames, we generate a set of object trajectory hypotheses with respect to semantic categories via object detection and temporal association which characterize the long-range spatio-temporal evolution of various object features and are commonly tied to higher-level contexts such as object interactions and behaviours \cite{WangC12,WangW14,WangHC14,liu2015multiclass,wang2016primary,wang2017a,wang2017,tang2018object}. 

  Specifically, we firstly extract generic object proposals by applying MCG \cite{arbelaez2014multiscale} in each frame. Object detection is performed on this pool of object proposals by using faster R-CNN \cite{Ren2015}, which is trained on 20 PASCAL VOC classes. A set of object hypotheses $\mathbb{D}$ are formed by keeping proposals with detection confidence exceeding a threshold (0.5).

Object trajectory hypotheses $\mathbb{T}$ \wrt each semantic class are generated by temporally associating a cohort of object hypotheses $\mathbb{D}$ by imposing frame-to-frame spatio-temporal consistency, similar to  \cite{wang2017}. Specifically, we utilize object tracker \cite{ma2015hierarchical} to track object hypotheses over time to both ends of the video sequence as follows.
\begin{itemize}
	\item Initialize an empty trajectory hypothesis $T_i \in \mathbb{T}$
	\item Rank remaining object hypotheses in $\mathbb{D}$ based on detection confidence
	\item Initialize tracker with the bounding box of the highest ranked object hypothesis and perform tracking to both directions simultaneously
	\item Select object hypothesis in the new frame which have a sufficient overlap, \ie Intersection-over-Union (IoU) higher than a threshold (0.5), with the tracker box which is added to $T_i$ and consequently removed from $\mathbb{D}$.
\end{itemize}
The above steps are iteratively performed until no new trajectory hypothesis containing three or more object instances can be generated from $\mathbb{D}$. Fig. \ref{fig:diagram} shows exemplars of object trajectory hypotheses extracted from a video sequence. Regions \cite{felzenszwalb2004efficient} are extracted from each frame as the atomic data units. Let $\mathcal{R}_D$ be the set of regions constituting video object hypotheses, and $\mathcal{R}_U$ be the unlabeled regions.

\subsection{Graph Construction}

We firstly initialize a k-nearest neighbor similarity graph $\mathcal{G}=(\mathcal{V},\mathcal{E})$ between all $N$ regions from  $\mathcal{R}_D \cup \mathcal{R}_U$. Each vertex $v_{i} \in \mathcal{V}$ of the graph is described by the L2-normalized VGG-16 Net \cite{vggnet}  \emph{fc6} features $f_i$ of the corresponding region. Each weight $w_{i,j} \in \mathbf{W}$ of edge $e_{i,j} \in \mathcal{E}$ is defined as the inner-product between the feature vectors of neighboring vertices, i.e., $w_{i,j} = <f_i, f_j>$.

\subsection{Context Modeling}
\label{sec:modeling}
Frames containing object hypotheses are considered as annotated data to generate context exemplars, as object trajectory hypotheses normally capture essential parts of video objects. Let $\mathcal{F}$ be this set of annotated frames and $\mathcal{\hat{F}}$ be all the other frames in current video sequence. A context exemplar consists of a pair of regions and the corresponding semantic labels. The intuition behind this setting is that one region with its semantic label supports the paired region to be labeled with its corresponding semantic label. This exemplar is able to encode the global interaction and co-occurrence of semantic objects beyond local spatial adjacencies. The goal is to impose the consistency between each pair of regions from un-annotated frames and the extracted context exemplars. 

Formally, given a set of semantic labels $\mathcal{C} = \{c_0, c_1, \dots, c_{L-1}\}$ comprising all classes in the annotated data, we represent the context exemplars for each class pair $(c_m, c_n)$ as
\begin{equation} 
\mathbf{A^{m,n}}  = \{(v_i, v_j): C(v_i) = c_m, C(v_j) = c_n, v_i, v_j \in \mathcal{F} \} \nonumber
\end{equation}
where $v_i, v_j \in \mathcal{F}$ stands for two regions $v_i$ and $v_j$ from the annotated frame set $\mathcal{F}$ and 
$C(v_i)$ represents the semantic label of region $v_i$. Hence, all object class pairs as well as contextual relationships in the annotated frames are represented as $\mathcal{A} = \{\mathbf{A}^{0,0}, \mathbf{A}^{0,1}, \dots, \mathbf{A}^{L-1,L-1}\}$.

We transform the above context exemplar to a context link view of contextual knowledge, where context exemplar $(v_i, v_j)$ can be referred to as a $(c_m, c_n)$-type link between two vertices on the similarity graph. Let $\mathcal{P}$ denote the set of $N\times N$ matrices, where a matrix $\mathbf{P}^{m,n}\in \mathcal{P}$ is associated with all $(c_m, c_n)$ class pair links.  Each entry $[\mathbf{P}^{m,n}]_{i,j} \in \mathbf{P}^{m,n}$ indicates the confidence of $(c_m, c_n)$-link between two regions $v_i$ and $v_j$. The confidence  ranging between 0 and 1 corresponds to the probability of the existence of a link, where 1 stands for high confidence of the existence of a link and 0 indicates the absence of a link. The $(c_m, c_n)$-links which have been observed within the annotated frames can be represented by another set of matrices $\mathbf{O}^{m,n}\in \mathcal{P}$ such that
\begin{equation} 
[\mathbf{O}^{m,n}]_{i,j} =
\left\{ \begin{array}{lll}
1 &  \mbox{if} & (v_i, v_j) \in \mathbf{A^{m,n}}  \\ 
0 & \mbox{otherwise} &
\end{array}\right.
\end{equation}
All the observed context link can be denoted as $\mathcal{O} = \{\mathbf{O}^{0,0}, \mathbf{O}^{0,1}, \dots, \mathbf{O}^{L-1,L-1}\}$.

\subsection{Context Prediction}
\label{sec:prediction}
Given the above context link view of contextual knowledge, 
we formulate the context prediction problem as a task of link prediction problem which determines how probable a certain link exists in a graph. To this end, 
we predict $(c_m, c_n)$-links among the pairs of vertices from $\mathcal{R}_U$ based on  $\mathbf{O}^{m,n}$ consistent to the intrinsic structure of the similarity
graph. Specifically, we propagate $(c_m, c_n)$-links in $\mathbf{O}^{m,n}$ to estimate the strength of the pairs of vertices from $\mathcal{R}_U$. We drop the $m,n$
suffix for clarity. 

\begin{algorithm}[t!]
	\caption{Context learning algorithm}\label{algo}
	\begin{algorithmic}[1]
		\Procedure{Link prediction}{}
		\State $S (v_i, c_i, v_j, c_j) \gets \varnothing$
		\State $\text{Graph} ~\mathcal{G} \gets \text{all}~\textit{regions of video}$
		\State $\text{Affinity matrix} ~\mathbf{W} \gets \textit{k-nearest neighbors} $
		\State $d_i=\sum_{j=1}^{N} w_{ij}$
		\State $\mathbf{D} \gets \mathrm{diag}([d_1, \dots, d_N])$
		\State $\mathbf{L} \gets \mathbf{D}^{-\frac{1}{2}} \mathbf{W}  \mathbf{D}^{-\frac{1}{2}}$
		\State $\mathcal{A} \gets \textit{context exemplars} $
		\State $\mathcal{O} \gets \textit{context links} \in \mathcal{A}$
		\For{\text{each} \textit{class pair $(c_m, c_n)$}}
		\State $\mathbf{P}_r(1)  \gets \mathbf{0}, ~ \mathbf{P}_c(1)  \gets \mathbf{0}$
		\State $\textit{Convergence} \gets \mathbf{false}$
		\While{$\textit{Convergence} ~\text{is}~ \mathbf{false}$} \Comment{row-wise}
		\State $\mathbf{P}_r(t+1) \gets \mu \mathbf{L} \mathbf{P}_r(t) + (1-\mu)\mathbf{O}^{m,n}$ 
		\EndWhile\label{euclidendwhile}
		\State $\textit{Convergence} \gets \mathbf{false}$
		\While{$\textit{Convergence} ~\text{is}~ \mathbf{false}$} \Comment{column-wise}
		\State $\mathbf{P}_c(t+1) \gets \mu \mathbf{L} \mathbf{P}_c(t) + (1-\mu)\mathbf{\hat{P}}_r$ 
		\EndWhile\label{euclidendwhile}	
		\State $S (v_i, c_i = c_m, v_j, c_j = c_n) \gets [\mathbf{\hat{P}}_c]_{ij}$
		\EndFor
		\EndProcedure
	\end{algorithmic}
\end{algorithm}

Directly solving the  link prediction problem is impractical for video segmentation since the complexity is as high as $O(N^4)$. Hence we 
propose to decompose the link propagation problem into two separate label propagation processes. As described in Algorithm \ref{algo},
row-wise link predication (step 13-14) is firstly performed, followed by column-wise link prediction (step 16-17). More specifically, the $j$-th row $\mathbf{O}^{j,.}$,
\ie the context exemplars associated with $v_j$, serves as an initial configuration of a label propagation problem \cite{Zhou2004} with respect to vertex $v_j$.
Each row is handled separately as a binary label propagation which converges to $\mathbf{\hat{P}}_r$. It is observed that the label propagation does not
apply to the rows of $\mathbf{O}$ corresponding to $\mathcal{R}_U$, and thus we only perform row-wise link propagation in rows corresponding to
annotated regions, which is much less than $N$. For the column-wise propagation, the $i$-th converged row $[\mathbf{\hat{P}}_r]_i$ is used 
to initialize the configuration. After convergence of the column-wise propagation, the probability of $(c_m, c_n)$-link between two vertices
of $\mathcal{R}_U$ is obtained. 

\subsection{Inference}
\label{sec:inference}
We formulate semantic video object segmentation as a region labeling problem, where the learned context link scores $S (v_i, c_i, v_j, c_j)$
can be incorporated while assigning labels to the set of regions $\mathcal{R}_D \cup \mathcal{R}_U$. We adopt the fully connected CRF 
that is proved to be effective in encoding model contextual relationships between object classes. 

Consider a random field $\mathbf{x}$ defined over a set of variables $\{x_0, \dots, x_{N-1}\}$, and the domain of each variable is a set of class
labels $\mathcal{C} = \{c_0, c_1, \dots, c_{L-1}\}$. The corresponding Gibbs energy is 
\begin{equation} 
E(\mathbf{x}) = \sum_{i} \psi(x_i) +  \sum_{i,j} \phi(x_i, x_j). \label{eq:graphcut}
\end{equation}

The unary potential $\psi(x_i)$ is defined as the negative logarithm of the likelihood of assigning $v_i$ with label $x_i$. 
To obtain $\psi(x_i)$, we learn a SVM 
model based on hierarchical CNN features \cite{ma2015hierarchical} by sampling from the 
annotated frames. 

The pairwise potential $\phi(x_i, x_j)$ encodes the contextual relationships between the regions learned via link prediction, which is defined 
as
\begin{equation} 
\phi(x_i, x_j) = \exp (-\frac{S (v_i, c_i, v_j, c_j)^2}{2\beta}) 
\end{equation}
where $\beta = <S (v_i, c_i, v_j, c_j)^2>$ is the adaptive weight and $<\cdot>$ indicates the expectation. 

We adopt a combined QPBO and $\alpha$-expansion inference (a.k.a fusion moves) \cite{lempitsky2010fusion} to optimize (\ref{eq:graphcut})  
and the resulting label assignment gives the semantic object segmentation of the video sequence. 

\section{Experiments}
\label{sec:evaluation}

We evaluate our proposed approach on YouTube-Objects \cite{PrestLCSF12}, which is the \emph{de facto} benchmark for assessing semantic video
object segmentation algorithms. The class labels of these two densely labeled datasets belong to the 20 classes of PASCAL VOC 2012.  The YouTube-Objects dataset consists of videos from $10$  classes with pixel-level ground truth for totally more than $20,000$ frames.  These videos are very challenging and completely unconstrained, with objects of similar colour to the background, fast motion, non-rigid deformations, and fast camera motion. We compare our approach with six state-of-the-art semantic video object segmentation methods which have been reported on this dataset, \ie  \cite{PrestLCSF12} (ODW),   \cite{TangSY013} (DSA), \cite{ZhangCLWX15} (SOD),  \cite{wang2016semi} (SDA),  \cite{drayer2016object} (DTS) and \cite{wang2017} (CGG). Standard average IoU is used to measure the segmentation accuracy, $IoU = \frac{S \cap G}{S \cup G}$, where $S$ is the segmentation result and $G$ stands for the ground-truth mask.

\begin{table}[t!]
	\caption{Intersection-over-union overlap on YouTube-Objects Dataset}
	\centering
	\begin{tabular}{lrrrrrrrrrrr}
		\toprule
		&   Aeroplane &   Bird &    Boat &  Car &  Cat &  Cow &  Dog & Horse & Mbike & Train & \shortstack{Avg.} \\
		\midrule
		ODW & 0.517  &  0.175 & 0.344  & 0.347   & 0.223 & 0.179 & 0.135 & 0.267 & 0.412 & 0.250 & 0.285\\
		DSA    &  0.178  & 0.198  & 0.225  & 0.383  & 0.236  & 0.268 & 0.237 & 0.140 & 0.125 & 0.404 & 0.239\\
		SOD   & 0.758   &  0.608  & 0.437  & 0.711  & 0.465 & 0.546 & 0.555 & 0.549 & 0.424 & 0.358 & 0.541\\
		SDA     &  0.760  &  0.747  & 0.588  & 0.659  & 0.557 & 0.675 & 0.574 & 0.575 & 0.569 & 0.430 & 0.613 \\
		DTS	  &  0.744  &  0.721 &  0.585  & 0.600   & 0.457 & 0.612 & 0.552 & 0.566 & 0.421 & 0.367 & 0.562\\
		CGG   &  0.757  &  0.766  & 0.666    & 0.758    &  0.624 & 0.720 & 0.671 & 0.526 & 0.547 & 0.392 & 0.643 \\
		\midrule
		Ours    &  \underline{\textbf{0.785}}  &  \underline{\textbf{0.772}}  &  \underline{\textbf{0.725}}  & \underline{\textbf{0.766}}  & \underline{\textbf{0.672}} & \underline{\textbf{0.731}} & \underline{\textbf{0.672}} &  \underline{\textbf{0.607}} & \underline{\textbf{0.614}} & \underline{\textbf{0.407}} & \underline{\textbf{0.675}} \\
		\bottomrule
	\end{tabular} \label{tbl:yto-result}
\end{table}

\begin{figure*}[t!]
	\centering
	\includegraphics[width=0.99\linewidth]{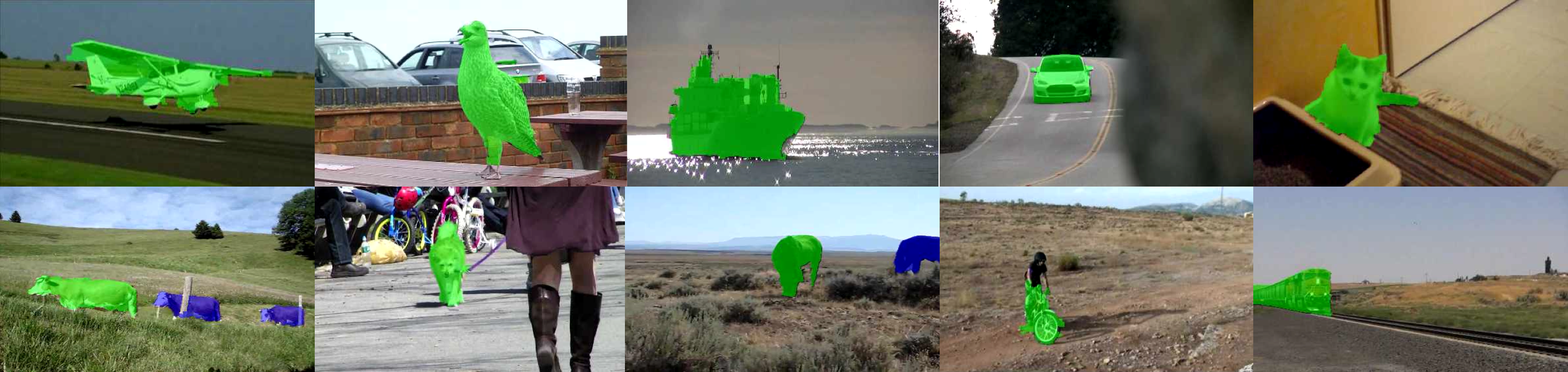}
	\caption{Qualitative results of our algorithm on YouTube-Objects Dataset.  \label{fig:yto}}
\end{figure*}

We summarize the comparisons of our algorithm with other approaches in Table \ref{tbl:yto-result}.  Table \ref{tbl:yto-result} demonstrates the superior performance of our proposed algorithm which surpasses the competing methods in all classes, with a significant increase of segmentation accuracy, \ie $3.2\%$ in average over the best competing method CGG. We attribute this improvement to the capability of learning and propagating higher-level spatial-temporal contextual relationships of video objects, as opposed to imposing contextual information in local labeling (CGG) or modeling local appearance (SDA).  One common limitation of  these methods is that they are error-prone in separating interacting objects exhibiting similar appearance or motion, which is intractable unless the inherent contextual relationship is explored.

Our algorithm outperforms another two methods which also utilize object detection, \ie DTS and SOD, with large margins of $11.3\%$ and $13.4\%$. DTS shares some similarity with our approach in that it also uses faster R-CNN for the initial object detection which makes it a comparable baseline to demonstrate the effectiveness of our algorithm. By exploiting contextual relationships in a global manner, our algorithm is able to account for object evolutions in the video data to resolve both appearance and motion ambiguities. SOD performs the worst among the three as it only conducts temporal association of detected object segments without explicitly modeling either the objects or contexts.  Some qualitative results of the proposed algorithm on YouTube-Objects dataset are shown in Fig. \ref{fig:yto}.

\section{Conclusion}

We have proposed a novel approach to modeling the semantic contextual relationships for tackling the challenging video object segmentation problem. The proposed model comprises an exemplar-based nonparametric view of contextual cues, which is formulated as link prediction problem solved by label propagation on a similarity graph of regions. The derived contextual relationships are utilized to estimate the pairwise contexts between all pairs of unlabeled local regions. The experiments demonstrated that modeling the semantic contextual relationships effectively improved
segmentation robustness and accuracy which significantly advanced the state-of-the-art on challenging benchmark. 
%
%
%
\bibliographystyle{splncs04}
\bibliography{refs}

\end{document}